\documentclass{article}

\PassOptionsToPackage{numbers, compress}{natbib}

\usepackage[final]{nips_2017}

\usepackage[utf8]{inputenc} 
\usepackage[T1]{fontenc}    
\usepackage{hyperref}       
\usepackage{url}            
\usepackage{booktabs}       
\usepackage{amsfonts}       
\usepackage{nicefrac}       

\usepackage{kotex}
\usepackage{amsmath}
\usepackage{graphicx}
\usepackage{algorithm}
\usepackage{algorithmic}

\usepackage{changepage}
\usepackage{array}
\newcolumntype{L}[1]{>{\raggedright\let\newline\\\arraybackslash\hspace{0pt}}m{#1}}
\newcolumntype{C}[1]{>{\centering\let\newline\\\arraybackslash\hspace{0pt}}m{#1}}

\usepackage{color}
\newcommand{\rt}{\textcolor[rgb]{0,0,0}}
\newcommand{\ot}{\textcolor[rgb]{0,0,0}}
\newcommand{\gt}{\textcolor[rgb]{0,0,0}}

\DeclareMathOperator*{\argminA}{argmin}

\title{Overcoming Catastrophic Forgetting by\\ Incremental Moment Matching}

\author{
  Sang-Woo Lee$^1$, Jin-Hwa Kim$^1$, Jaehyun Jun$^1$, Jung-Woo Ha$^2$, 
  and Byoung-Tak Zhang$^{1,3}$\\ 
  \\
  Seoul National University$^1$\\
  Clova AI Research, NAVER Corp$^2$\\
  Surromind Robotics$^3$\\
  \\
  \texttt{\{slee,jhkim,jhjun\}@bi.snu.ac.kr }
  \texttt{jungwoo.ha@navercorp.com}\\
  \texttt{btzhang@bi.snu.ac.kr}
}

\begin{document}

\maketitle

\begin{abstract}
Catastrophic forgetting is a problem of neural networks that loses the information of the first task after training the second task.
Here, we propose a method, i.e. incremental moment matching (IMM), to resolve this problem. 
IMM incrementally matches the moment of the posterior distribution of the neural network which is trained on the first and the second task, respectively.
To make the search space of posterior parameter smooth, the IMM procedure is complemented by various transfer learning techniques including weight transfer, L2-norm of the old and the new parameter, and a variant of dropout with the old parameter.
We analyze our approach on a variety of datasets including the MNIST, CIFAR-10, Caltech-UCSD-Birds, and Lifelog datasets.
The experimental results show that IMM achieves state-of-the-art performance by balancing the information between an old and a new network.
\end{abstract}


\section{Introduction}

Catastrophic forgetting is a fundamental challenge for artificial general intelligence based on neural networks.
The models that use stochastic gradient descent often forget the information of previous tasks after being trained on a new task \cite{mccloskey1989,french1999}.
Online multi-task learning that handles such problems is described as \textit{continual learning}.
This classic problem has resurfaced with the renaissance of deep learning research \cite{goodfellow2013,srivastava2013}.

Recently, the concept of applying a regularization function to a network trained by the old task to learning a new task has received much attention.
This approach can be interpreted as an approximation of sequential Bayesian \cite{Ghahramani2000,Broderick2013}.
Representative examples of this regularization approach include learning without forgetting \cite{li2016} and elastic weight consolidation \cite{kirkpatrick2016}.
These algorithms succeeded in some experiments where their own assumption of the regularization function fits the problem.

Here, we propose incremental moment matching (IMM) to resolve the catastrophic forgetting problem.
IMM uses the framework of Bayesian neural networks, which implies that uncertainty is introduced on the parameters in neural networks, and that the posterior distribution is calculated \cite{mackay1992,Blundell2015}. The dimension of the random variable in the posterior distribution is the number of the parameters in the neural networks.
IMM approximates the mixture of Gaussian posterior with each component representing parameters for a single task to one Gaussian distribution for a combined task.
To merge the posteriors, we introduce two novel methods of moment matching.
One is \textit{mean-IMM}, which simply averages the parameters of two networks for old and new tasks as the minimization of \rt{the average of KL-divergence between one approximated posterior distribution for the combined task and each Gaussian posterior for the single task \cite{goldberger2005}.} The other is \textit{mode-IMM}, which merges the parameters of two networks using a Laplacian approximation \cite{mackay1992} to approximate a mode of the mixture of two Gaussian posteriors, which represent the parameters of the two networks.

\begin{figure}[t] 
\centering
\includegraphics[width=0.80\textwidth]{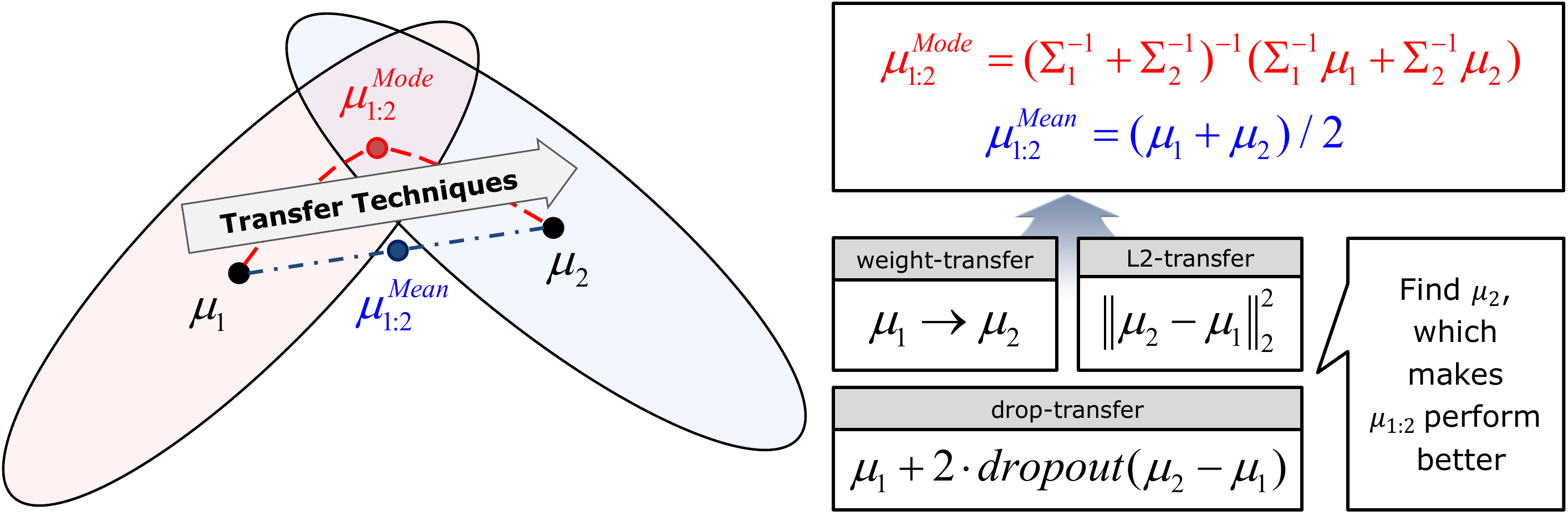}
\caption{Geometric illustration of incremental moment matching (IMM). Mean-IMM simply averages the parameter\rt{s} of two neural networks, whereas mode-IMM \ot{tries to find} a maximum of the mixture of Gaussian posteriors.
\rt{To make IMM be} reasonable, the search space of the loss function between the posterior means $\mu_1$ and $\mu_2$ should be \ot{reasonably} smooth and convex-like. 
To find a $\mu_2$ which satisfies this condition of a smooth and convex-like path from $\mu_1$, we propose \ot{applying} various transfer techniques for the IMM procedure.
}
\label{fig:fig1}
\end{figure}

In general, it is too na{\"i}ve to assume that the final posterior distribution for the whole task is Gaussian.
To make our IMM work, the search space of the loss function between the posterior means needs to be smooth and convex-like.
In other words, there should not be high cost barriers between the means of the two networks for an old and a new task.
To make our assumption of Gaussian distribution for neural network reasonable, we applied three main transfer learning techniques on the IMM procedure: weight transfer, L2-norm of the old and the new parameters, and our newly proposed variant of dropout using the old parameters.
The whole procedure of IMM is illustrated in Figure \ref{fig:fig1}.



\section{Previous Works on Catastrophic Forgetting}

One of the major approaches preventing catastrophic forgetting is to use an ensemble of neural networks. 
When a new task arrives, the algorithm makes a new network, and shares the representation between the tasks \cite{lee2016,rusu2016}.
However, this approach has a complexity issue, especially in inference, because the number of networks increases \rt{as} the number of new tasks that need to be learned \rt{increases.}

Another approach studies \rt{the} methods using implicit distributed storage of information\ot{, in typical} stochastic gradient descent (SGD) learning.
These methods use the idea of dropout, maxout, or neural module \rt{to} distributively \rt{store} the information for each task by making use of the large capacity of the neural network \cite{srivastava2013}.
Unfortunately, most studies following this approach had limited success and failed to preserve performance on the old task when an extreme change to the environment occurred \cite{goodfellow2013}.
Alternatively, Fernando et al. \cite{fernando2017} proposed PathNet, which extends the idea of the ensemble approach for parameter reuse \cite{rusu2016} within a single network.
In PathNet, a neural network has ten or twenty modules in each layer, and three or four modules are picked for one task in each layer by an evolutionary approach.
This method alleviates the complexity issue of the ensemble approach to continual learning in a
plausible way.

The approach with \rt{a} regularization term also \rt{has} \rt{received attention}.
Learning without forgetting (LwF) is one example of this approach, which uses the pseudo-training data from the old task \cite{li2016}.
Before learning the new task, LwF puts the training data of the new task into the old network, and uses the output as pseudo-labels of the pseudo-training data.
By optimizing both the pseudo-training data of the old task and the real data of the new task, LwF attempts to prevent catastrophic forgetting.
This framework is promising where the properties of the pseudo training set is similar to the ideal training set. 
Elastic weight consolidation (EWC), another example of this approach, uses sequential Bayesian estimation to update neural networks for continual learning \cite{kirkpatrick2016}.
In EWC, the posterior distribution trained by the previous task is used to update the new prior distribution.
This new prior is used for learning the new posterior distribution of the new task in a Bayesian manner.
EWC assumes that the covariance matrix of the posterior is diagonal and there are no correlations between the nodes. Though this assumption is \rt{fragile}, EWC performs well in some domains.

\gt{EWC is a monumental recent work that uses sequential Bayesian for continual learning of neural networks.
However, updating the parameter of complex hierarchical models by sequential Bayesian estimation is not new \cite{Ghahramani2000}. Sequential Bayes was used to learn topic models from stream data} \rt{by} \gt{Broderick et al. \cite{Broderick2013}.}
\rt{Huang et al. applied sequential Bayesian to adapt a deep neural network to the specific user in the speech recognition domain \cite{huang2014,huang2015}.
They assigned the layer for the user adaptation and applied MAP estimation to this single layer.}
\gt{Similar to our IMM method, Bayesian moment matching is used for sum-product networks, a kind of deep hierarchical probabilistic model \cite{rashwan2016}.
Though sum-product networks are usually not scalable to large datasets, their online learning method is useful, and it achieves similar performance to the batch learner. 
Our method using moment matching focuses on continual learning and deals with significantly different statistics between tasks, unlike the previous method.} 


\section{Incremental Moment Matching}


In incremental moment matching (IMM), the moments of posterior distributions are matched in an incremental way.
In our work, we use a Gaussian distribution to approximate the posterior distribution of parameters.
Given $K$ sequential tasks, we want to find the optimal parameter $\mu^*_{1:K}$ and $\Sigma^*_{1:K}$ of the Gaussian approximation function $q_{1:K}$ from the posterior parameter for each $k$th task, $(\mu_k,\Sigma_k)$.

\begin{gather}
    p_{1:K} \equiv p(\theta|X_1,\cdots,X_K,y_1,\cdots,y_K) \approx q_{1:K} \equiv q(\theta|\mu_{1:K},\Sigma_{1:K})
    \label{eq:eq1}\\
    p_k \equiv p(\theta|X_k,y_k) \approx q_k \equiv q(\theta|\mu_k,\Sigma_k)
    \label{eq:eq2}
\end{gather}

\ot{$q_{1:K}$} denotes an approximation of the true posterior distribution $p_{1:K}$ for the whole task, and $q_k$ denotes an approximation of the true posterior distribution $p_k$ over the training dataset $(X_k,y_k)$ for the $k$th task. $\theta$ denotes the vectorized parameter of the neural network. 
The dimension of $\mu_k$ and $\mu_{1:k}$ is $D$, and the dimension of $\Sigma_k$ and $\Sigma_{1:k}$ is $D \times D$, respectively, where $D$ is the dimension of $\theta$.
For \rt{example, a} multi-layer perceptrons (MLP) with \rt{[784-800-800-800-10] has} the number of nodes, $D$ = 1917610 \rt{including bias terms.}

Next, we explain two proposed moment matching algorithms for the continual learning of \ot{modern} deep neural networks. \rt{The two algorithms generate two different moments of Gaussian with different objective functions} for the same dataset.

\subsection{Mean-based Incremental Moment Matching (mean-IMM)}
\textbf{Mean-IMM} averages the parameters of two networks in each layer, using mixing ratios $\alpha_k$ with $\sum^K_k \alpha_k = 1$.
The objective function of mean-IMM is to minimize 
\rt{the following \textit{local KL-distance} or the weighted sum of KL-divergence between each $q_k$ and $q_{1:K}$ \cite{goldberger2005,zhang10}:}

\begin{gather}
    \mu^*_{1:K}, \Sigma^*_{1:K} = \argminA_{\mu_{1:K},\Sigma_{1:K}} 
    \rt{\mbox{$\sum^K_k$} \alpha_k KL ( q_k || q_{1:K} )} \label{eq:eq4}\\
    \mu^*_{1:K} = \mbox{$\sum^K_k$} \alpha_k \mu_k \label{eq:eq5}\\
    \Sigma^*_{1:K} = \mbox{$\sum^K_k$} \alpha_k ( \Sigma_k + (\mu_k-\mu^*_{1:K})(\mu_k-\mu^*_{1:K})^T)
\end{gather}

\rt{$\mu^*_{1:K}$ and $\Sigma^*_{1:K}$ are the optimal solution of the local KL-distance.}
Notice that covariance information is not needed for mean-IMM, since calculating $\mu^*_{1:K}$ does not require any $\Sigma_k$. A series of $\mu_k$ is sufficient to perform the task.
The idea of mean-IMM is commonly used in shallow networks \cite{Pathak10,Baldi13}. 
However, the contribution of this paper is to discover \rt{when} and how mean-IMM can be applied in modern deep neural networks and to show it can performs better with other transfer techniques. 

\rt{Future works may include other measures to merge the networks, including the KL-divergence between $q_{1:K}$ and the mixture of each $q_k$ (i.e. $KL ( q_{1:K}||\mbox{$\sum^K_k$} \alpha_k q_k )$) \cite{zhang10}.}
\subsection{Mode-based Incremental Moment Matching (mode-IMM)}

\textbf{Mode-IMM} is a variant of mean-IMM which uses the covariance information of the posterior of Gaussian distribution.
\rt{In general,} a weighted average of two mean vectors of Gaussian distributions is not a mode of MoG.
In discriminative learning, the maximum of the distribution is of primary interest.
According to Ray and Lindsay \cite{ray2005}, all the modes of MoG with $K$ clusters lie on $(K-1)$-dimensional hypersurface $\{ \theta | \theta = (\sum^K_k a_k \Sigma^{-1}_k )^{-1} (\sum^K_k a_k \Sigma^{-1}_k \mu_k), 0 < a_k < 1$ and $\sum_k a_k = 1 \}$.
See Appendix A for more detail\rt{s}.

Motivated by the above description, a mode-IMM approximate MoG with Laplacian approximation, in which the logarithm of the function is expressed by \rt{the} \ot{Taylor expansion} \cite{mackay1992}.
Using \ot{Laplacian approximation}, the MoG is approximated as follows:

\begin{gather}
    \log q_{1:K} \approx \mbox{$\sum^K_k$} \alpha_k \log q_k + C =
    -\frac{1}{2} \theta^T (\mbox{$\sum^K_k$} \alpha_k \Sigma^{-1}_k ) \theta + (\mbox{$\sum^K_k$} \alpha_k \Sigma^{-1}_k \mu_k) \theta + C'
\end{gather}
\begin{gather}
    \mu^*_{1:K} = \Sigma^{*}_{1:K} \cdot (\mbox{$\sum^K_k$} \alpha_k \Sigma^{-1}_k \mu_k)
    \label{eq:eq9}\\
    \Sigma^{*}_{1:K} = (\mbox{$\sum^K_k$} \alpha_k \Sigma^{-1}_k)^{-1}
    \label{eq:eq10}
\end{gather}

\rt{For Equation \ref{eq:eq10}, we add $\epsilon I$ to the term to be inverted in practice, with an identity matrix $I$ and a small constant $\epsilon$.}

Here, we assume diagonal covariance matrices, which means that there is no correlation among parameters. 
This diagonal assumption is useful, since it decreases the number of parameters for each covariance matrix from $O(D^2)$ to $O(D)$ for the dimension of the parameters $D$.

For covariance, we use the inverse of a Fisher information matrix, following \cite{kirkpatrick2016,pascanu2013}. 
The main idea of this approximation is that the square of gradients for parameters is a good indicator of their precision, which is the inverse of the variance. 
The Fisher information matrix for the $k$th task, $F_k$ is defined as:

\begin{equation}
\begin{aligned}
    F_k = E \left[\frac{\partial}{\partial \mu_k} \ln p(\tilde{y}|x,\mu_k) \cdot
    \frac{\partial}{\partial \mu_k} \ln p(\tilde{y}|x,\mu_k)^T \right],
\end{aligned}
\end{equation}

where the probability of the expectation follows $x \sim \pi_k$ and $\tilde{y} \sim p(y|x,\mu_k)$, where $\pi_k$ denotes an empirical distribution of $X_k$.


\section{Transfer Techniques for Incremental Moment Matching}
\label{sec:transfer_techniques}

In general, the loss function of neural networks is not convex. 
\rt{Consider that shuffling nodes and their weights in a neural network preserves the original performance.}
If the parameters of two neural networks initialized independently are averaged, it might perform poorly 
\rt{because of the} high cost barriers between the \rt{parameters of the two neural networks} \cite{goodfellow2014}.
However, we will show that various transfer learning techniques can be used to ease this problem, and make the assumption of Gaussian distribution for neural networks reasonable. In this section, we introduce three practical techniques for IMM, including weight-transfer, L2-transfer, and drop-transfer.

\subsection{Weight-Transfer}

\textbf{Weight-transfer} initialize the parameters for the new task $\mu_k$ with the parameters of the previous task $\mu_{k-1}$ \cite{yosinski2014}.
In our experiments, the use of weight-transfer was critical to the continual learning performance. For this reason, the experiments on IMM in this paper use the weight-transfer technique \rt{by} default.

The weight-transfer technique is motivated by the geometrical property of neural networks discovered in the previous work \cite{goodfellow2014}. They found that there is a straight path from the initial point to the solution without any high cost barrier, in various types of neural networks and datasets. This discovery suggests that the weight-transfer from the previous task to the new task makes a smooth loss surface between two solutions for the tasks, so that the optimal solution for both tasks lies on the interpolated point of the two solutions.

\begin{figure}[t] 
\includegraphics[width=\textwidth]{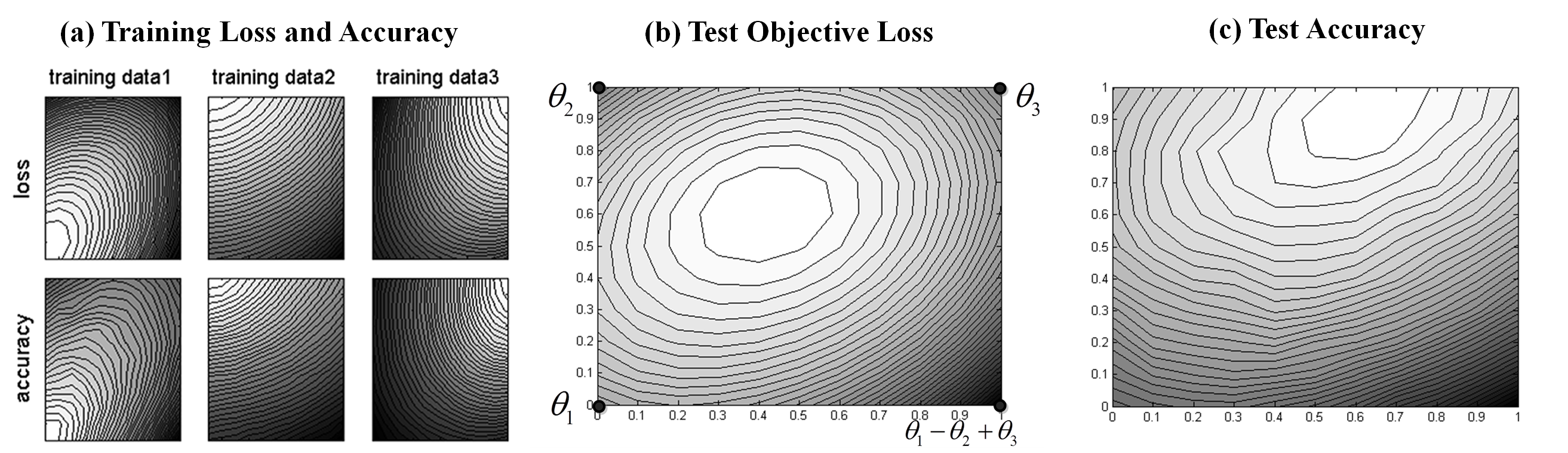}
\vskip -0.00in
 \caption{Experimental results on visualizing the effect of weight-transfer. The geometric property of the parameter space of the neural network is analyzed. Brighter is better.
 $\theta_1$, $\theta_2$, and $\theta_3$ are the \rt{vectorized parameters} of trained networks from randomly selected subsets of the CIFAR-10 dataset.
 This figure shows that there are better solutions between the three locally optimized parameters.}
\label{fig:fig2}
\end{figure}

To empirically validate the concept of weight-transfer, we use the linear path analysis proposed by Goodfellow et al. \cite{goodfellow2014} (Figure \ref{fig:fig2}). 
We randomly chose 18,000 instances from the training dataset of CIFAR-10, and divided them into three subsets of 6,000 instances each. 
These three subsets are used for sequential training by CNN models, parameterized by $\theta_1$, $\theta_2$, and $\theta_3$, respectively. Here, $\theta_2$ is initialized from $\theta_1$, and then $\theta_3$ is initialized from $\theta_2$, in the same way as weight-transfer.
In \rt{this} analysis, each loss and accuracy is evaluated at a series of points $\theta = \theta_1 + \alpha(\theta_2-\theta_1)+\beta(\theta_3-\theta_2)$, varying $\alpha$ and $\beta$.
In Figure \ref{fig:fig2}, the loss surface of the model on each online subset is nearly convex.
The figure shows that the parameter at $\frac{1}{3}(\theta_1 + \theta_2 + \theta_3)$, which is the same as the solution of mean-IMM, performs better than any other reference points $\theta_1$, $\theta_2$, or $\theta_3$.
However, when $\theta_2$ is not initialized by $\theta_1$, \ot{the convex-like shape disappears,} since there is a high cost barrier between the loss function of $\theta_1$ and $\theta_2$.

\subsection{L2-transfer}

\textbf{L2-transfer} is a variant of L2-regularization.
L2-transfer can be interpreted as a special case of EWC where the prior distribution is Gaussian with $\lambda I$ as a covariance matrix.
In L2-transfer, a regularization term of the distance between $\mu_{k-1}$ and $\mu_k$ is added to the following objective function for finding $\mu_k$, where $\lambda$ is a hyperparameter:

\begin{equation}
    \log p(y_k|X_k,\mu_k) - \lambda \cdot ||\mu_k - \mu_{k-1}||^2_2
    \label{eq:eq16}
\end{equation}

The concept of L2-transfer is commonly used in transfer learning \cite{evgeniou2004,kienzle2006}  and continual learning \cite{li2016,kirkpatrick2016} with large $\lambda$.
Unlike the previous usage of large $\lambda$, we use small $\lambda$ for the IMM procedure.
In other words, $\mu_k$ is first trained by Equation \ref{eq:eq16} with small $\lambda$, and then merged to $\mu_{1:k}$ in our IMM.
Since we want to make the loss surface between $\mu_{k-1}$ and $\mu_k$ smooth, and not to minimize the distance between $\mu_{k-1}$ and $\mu_k$.
In convex optimization, the L2-regularizer makes the convex function \ot{strictly convex}.
Similarly, we \ot{hope} L2-transfer with small $\lambda$ \rt{help to} find a $\mu_k$ with a convex-like loss space between $\mu_{k-1}$ and $\mu_k$.

\subsection{Drop-transfer}

\textbf{Drop-transfer} is a novel method devised in this paper.
\ot{Drop-transfer is a variant of dropout} where $\mu_{k-1}$ is the zero point of the dropout procedure.
In the training phase, the following $\hat{\mu}_{k,i}$ is used for the weight vector corresponding to the $i$th node $\mu_{k,i}$:

\begin{equation}
\label{eq:eq17}
\hat{\mu}_{k,i}=
\begin{cases}
 \mu_{k-1,i}, & \mbox{if } i \mbox{th node is turned off} \\
 \frac{1}{1-p} \cdot \mu_{k,i} - \frac{p}{1-p} \cdot \mu_{k-1,i}, & \mbox{otherwise}
\end{cases}
\end{equation}

where $p$ is the dropout ratio. Notice that the expectation of $\hat{\mu}_{k,i}$ is $\mu_{k,i}$.

\begin{table}[t]
  \centering
  \scriptsize	

  \caption{The averaged accuracies on the disjoint MNIST for two sequential tasks (Top) and the shuffled MNIST for three sequential tasks (Bottom). The untuned setting refers to \ot{the most natural} hyperparameter in the equation of each algorithm, whereas the tuned setting refers to using heuristic hand-tuned hyperparameters. 
  Hyperparam denotes the main hyperparameter of each algorithm. 
  For IMM with transfer, only $\alpha$ is tuned.
  The numbers in the parentheses refer to standard deviation. Every IMM uses weight-transfer.}
    
  \begin{tabular}{ l c c c c c}
    \hline
     & Explanation of  & \multicolumn{2}{c}{Untuned} & \multicolumn{2}{c}{Tuned}  \\
    \textbf{Disjoint MNIST Experiment}  & Hyperparam  & Hyperparam & Accuracy & Hyperparam & Accuracy \\ \hline
    
    SGD \cite{goodfellow2013} & epoch per dataset & 10 & 47.72 ($\pm$ 0.11)
    & 0.05  & 71.32 ($\pm$ 1.54)  \\
    L2-transfer \cite{evgeniou2004} & $\lambda$ in \eqref{eq:eq16} & - & - & 0.05 & 85.81 ($\pm$ 0.52) \\
    \textbf{Drop-transfer} & $p$ in \eqref{eq:eq17} & 0.5 & 51.72 ($\pm$ 0.79) & 0.5 & 51.72 ($\pm$ 0.79) \\
    EWC \cite{kirkpatrick2016} & $\lambda$ in \eqref{eq:eq15} & 1.0 & 47.84 ($\pm$ 0.04)
    & 600M & 52.72 ($\pm$ 1.36)  
    \\ \hline

    \textbf{Mean-IMM} & $\alpha_2$ in \eqref{eq:eq5} & 0.50 & 90.45 ($\pm$ 2.24)
    & 0.55 & 91.92  ($\pm$ 0.98) \\
    \textbf{Mode-IMM} & $\alpha_2$ in \eqref{eq:eq9} & 0.50 & 91.49 ($\pm$ 0.98)
    & 0.45 & 92.02  ($\pm$ 0.73)    
    \\ \hline
    
    \textbf{L2-transfer + Mean-IMM} & $\lambda$ / $\alpha_2$  & 0.001 / 0.50 & 78.34   ($\pm$ 1.82)
    & 0.001 / 0.60 & 92.62  ($\pm$ 0.95)  \\
    \textbf{L2-transfer + Mode-IMM} & $\lambda$ / $\alpha_2$ & 0.001 / 0.50 & 92.52  ($\pm$ 0.41)
    & 0.001 / 0.45 & 92.73  ($\pm$ 0.35) \\ \hline

    \textbf{Drop-transfer + Mean-IMM} & $p$ / $\alpha_2$ & 0.5 / 0.50 & 80.75  ($\pm$ 1.28)
    & 0.5 / 0.60 & 92.64 ($\pm$ 0.60)  \\
    \textbf{Drop-transfer + Mode-IMM} & $p$ / $\alpha_2$ & 0.5 / 0.50 & 93.35 ($\pm$ 0.49)
    & 0.5 / 0.50 & 93.35 ($\pm$ 0.49) \\ \hline

    \textbf{L2, Drop-transfer + Mean-IMM} & $\lambda$ / $p$ / $\alpha_2$ & 0.001 / 0.5 / 0.50 & 66.10 ($\pm$ 3.19) & 0.001 / 0.5 / 0.75 & \textbf{93.97} ($\pm$ 0.23)  \\
    \textbf{L2, Drop-transfer + Mode-IMM} & $\lambda$ / $p$ / $\alpha_2$ & 0.001 / 0.5 / 0.50 & \textbf{93.97} ($\pm$ 0.32) & 0.001 / 0.5 / 0.45 & \textbf{94.12} ($\pm$ 0.27)  \\ \hline 
    \\
    \textbf{Shuffled MNIST Experiment} & & Hyperparam & Accuracy & Hyperparam & Accuracy \\ \hline
    
    SGD \cite{goodfellow2013} & epoch per dataset & 60 & 89.15 ($\pm$ 2.34)  & - & $\sim$95.5 \cite{kirkpatrick2016}  \\
    L2-transfer \cite{evgeniou2004} & $\lambda$ in \eqref{eq:eq16} & - & - & 1e-3 & 96.37 ($\pm$ 0.62)  \\
    \textbf{Drop-transfer} & $p$ in \eqref{eq:eq17} & 0.5 & 94.75 ($\pm$ 0.62) & 0.2 & 96.86 ($\pm$ 0.21) \\
    EWC \cite{kirkpatrick2016} & $\lambda$ in \eqref{eq:eq15} & - & - & - & \textbf{$\sim$98.2} \cite{kirkpatrick2016} \\ \hline

    \textbf{Mean-IMM} & $\alpha_3$ in \eqref{eq:eq5} & 0.33 & 93.23 ($\pm$ 1.37) & 0.55 & 95.02  ($\pm$ 0.42) \\ 
    \textbf{Mode-IMM} & $\alpha_3$ in \eqref{eq:eq9} & 0.33 & 98.02 ($\pm$ 0.05) & 0.60 & 98.08 ($\pm$ 0.08) \\ \hline

    \textbf{L2-transfer + Mean-IMM} & $\lambda$ / $\alpha_3$  & 1e-4 / 0.33 & 90.38 ($\pm$ 1.74) & 1e-4 / 0.65 & 95.93 ($\pm$ 0.31)  \\
    \textbf{L2-transfer + Mode-IMM} & $\lambda$ / $\alpha_3$ & 1e-4 / 0.33 & \textbf{98.16 ($\pm$ 0.08)} & 1e-4 / 0.60 & \textbf{98.30 ($\pm$ 0.08)}  \\ \hline

    \textbf{Drop-transfer + Mean-IMM} & $p$ / $\alpha_3$ & 0.5 / 0.33 & 90.79 ($\pm$ 1.30) & 0.5 / 0.65  & 96.49 ($\pm$ 0.44)  \\
    \textbf{Drop-transfer + Mode-IMM} & $p$ / $\alpha_3$ & 0.5 / 0.33 & 97.80 ($\pm$ 0.07) & 0.5 / 0.55 & 97.95 ($\pm$ 0.08)  \\ \hline

    \textbf{L2, Drop-transfer + Mean-IMM} & $\lambda$ / $p$ / $\alpha_3$ & 1e-4 / 0.5 / 0.33  & 89.51 ($\pm$ 2.85)   & 1e-4 / 0.5 / 0.90 & 97.36 ($\pm$ 0.19)  \\
    \textbf{L2, Drop-transfer + Mode-IMM} & $\lambda$ / $p$ / $\alpha_3$ & 1e-4 / 0.5 / 0.33 & 97.83 ($\pm$ 0.10) & 1e-4 / 0.5 / 0.50 & 97.92 ($\pm$ 0.05) \\ \hline
  \end{tabular}

  \label{table:table1}
\end{table}

There are studies \cite{srivastava2014, Baldi13} that have interpreted dropout as an exponential ensemble of weak learners. 
By this perspective, \rt{since} the marginalization of output distribution over the whole weak learner is intractable, the parameters multiplied by the inverse of the dropout rate are used for the test phase in the procedure. In other words, the parameters of the weak learners are, in effect, simply averaged oversampled learners by dropout.
At the process of drop-transfer in our continual learning setting, we hypothesize that the dropout process makes the averaged point of two arbitrary sampled points using Equation~\ref{eq:eq17} a good estimator.

We investigated the search space of the loss function of the MLP trained from the MNIST handwritten digit recognition dataset for with and without dropout regularization, to supplement the evidence of the described hypothesis.
Dropout regularization makes the accuracy of a sampled point from dropout distribution and an average point of \rt{two sampled parameters}, from 0.450 ($\pm$ 0.084) to 0.950 ($\pm$ 0.009) and 0.757 ($\pm$ 0.065) to 0.974 ($\pm$ 0.003), respectively.
For the case of both with and without dropout, the space between two arbitrary samples is empirically convex, and fits \rt{to} the second-order equation.
Based on this experiment, we expect not only that the search space of the loss function between \ot{modern} neural networks \rt{can be easily} nearly convex \cite{goodfellow2014}, but also that regularizers, such as dropout, make the search space smooth and the point in the search space \rt{have} a good accuracy in continual learning. 




\section{Experimental Results}

We evaluate our approach on four experiments, \rt{whose} settings are intensively used in the previous works \cite{srivastava2013,kirkpatrick2016,li2016,lee2016}. 
For more details and experimental results, see Appendix D.
The source code for the experiments is available in Github repository\footnote{https://github.com/btjhjeon/IMM\_tensorflow}.

\textbf{Disjoint MNIST Experiment. }
The first experiment is the disjoint MNIST experiment \cite{srivastava2013}.
In this experiment, the MNIST dataset is divided into two datasets: the first dataset consists of only digits \{0, 1, 2, 3, 4\} and the second dataset consists of the remaining digits \{5, 6, 7, 8, 9\}.
Our task is 10-class joint categorization, unlike \rt{the setting} in the previous work, 
\ot{which considers two independent tasks of 5-class categorization.}
Because the inference should \rt{decide} whether a new instance comes from the first or the second task, our task is more difficult than the task \rt{of} the previous work.

\begin{figure}[t] 
\centering
\begin{adjustwidth}{-0.1cm}{-0.1cm}
\includegraphics[width=0.333\textwidth,trim={0.4cm 0 0.4cm 0},clip]{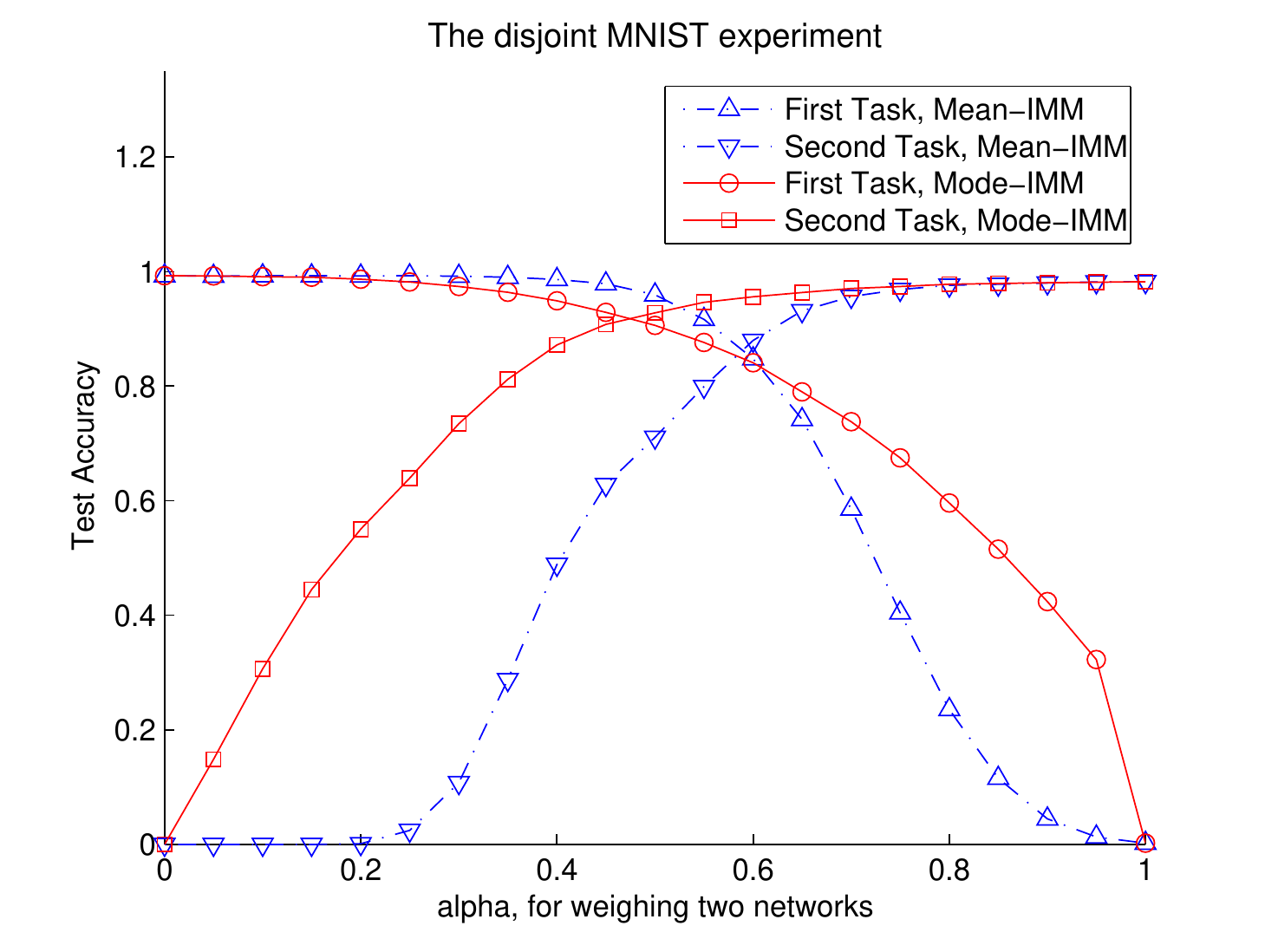}
\includegraphics[width=0.333\textwidth,trim={0.4cm 0 0.4cm 0},clip]{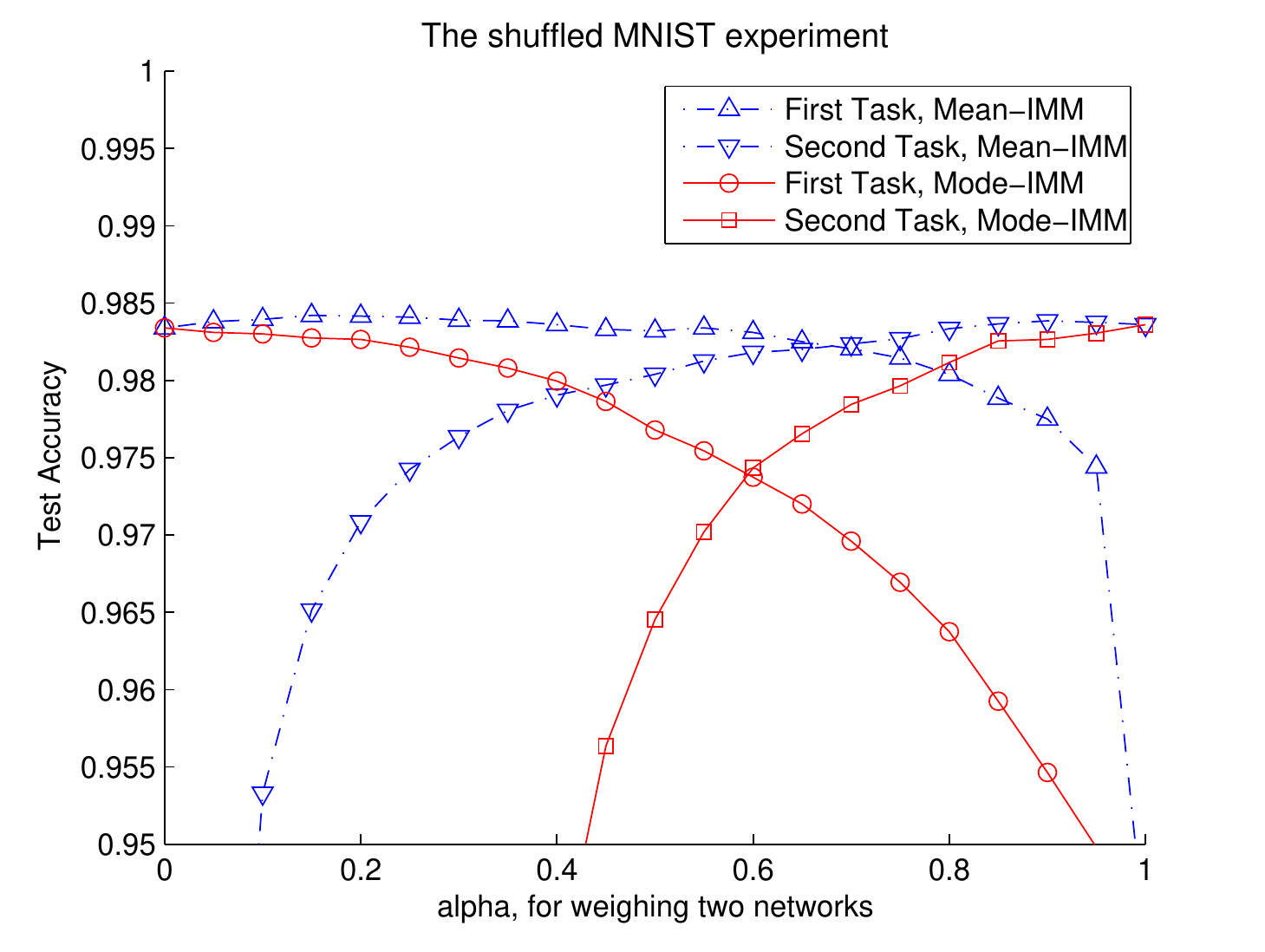}
\includegraphics[width=0.333\textwidth,trim={0.4cm 0 0.4cm 0},clip]{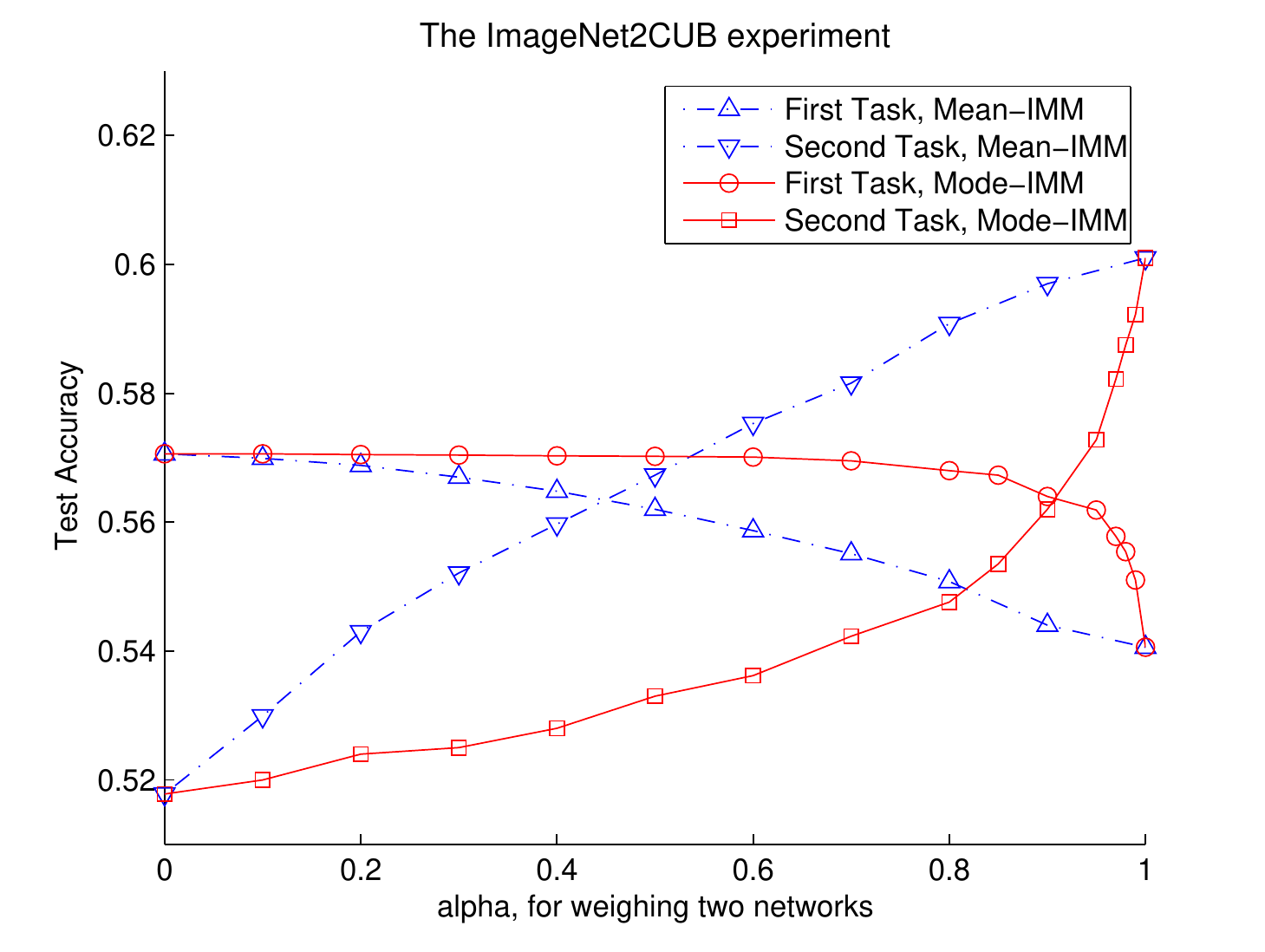}
\end{adjustwidth}
\caption{Test accuracies of two IMM models with weight-transfer on the disjoint MNIST (Left), the shuffled MNIST (Middle), and the ImageNet2CUB experiment (Right). $\alpha$ is a hyperparameter that balances the information between the old and the new task.}
\label{fig:fig3}
\end{figure}

We evaluate the models both on the untuned setting and the tuned setting.
The untuned setting refers to \ot{the most natural} hyperparameter in the equation of each algorithm.
The tuned setting refers to using heuristic hand-tuned hyperparameters.
Consider that tuned \rt{hyperparameter setting} is often used in previous works of continual learning as it is difficult to define a validation set in their setting. For example, when the model needs to learn from the new task after learning from the old task, a low learning rate or early stopping without a validation set, or arbitrary hyperparameter for balancing is used \cite{goodfellow2013,kirkpatrick2016}.
We discover hyperparameters in the tuned setting not only \rt{to find} the oracle performance of each algorithm, but also \rt{to show} that there \ot{exist} some paths consisting of the point that performs reasonably for both tasks.
Hyperparam in Table \ref{table:table1} denotes hyperparameter mainly searched in the tuned setting.
Table \ref{table:table1} (Top) and Figure \ref{fig:fig3} (Left) shows the experimental results from the disjoint MNIST experiment.

\begin{figure}[t] 
\centering
\includegraphics[width=0.40\textwidth]{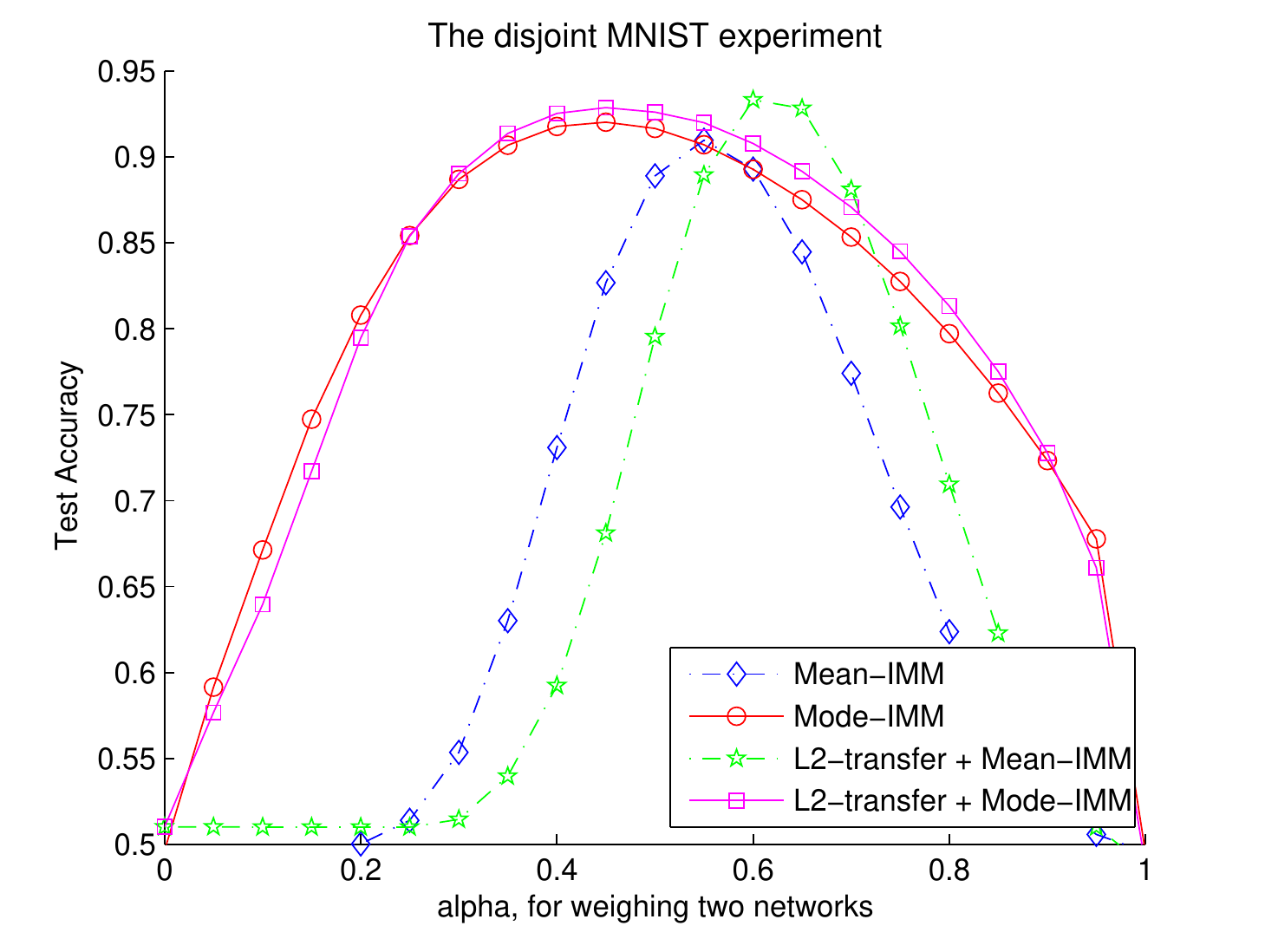}
\includegraphics[width=0.40\textwidth]{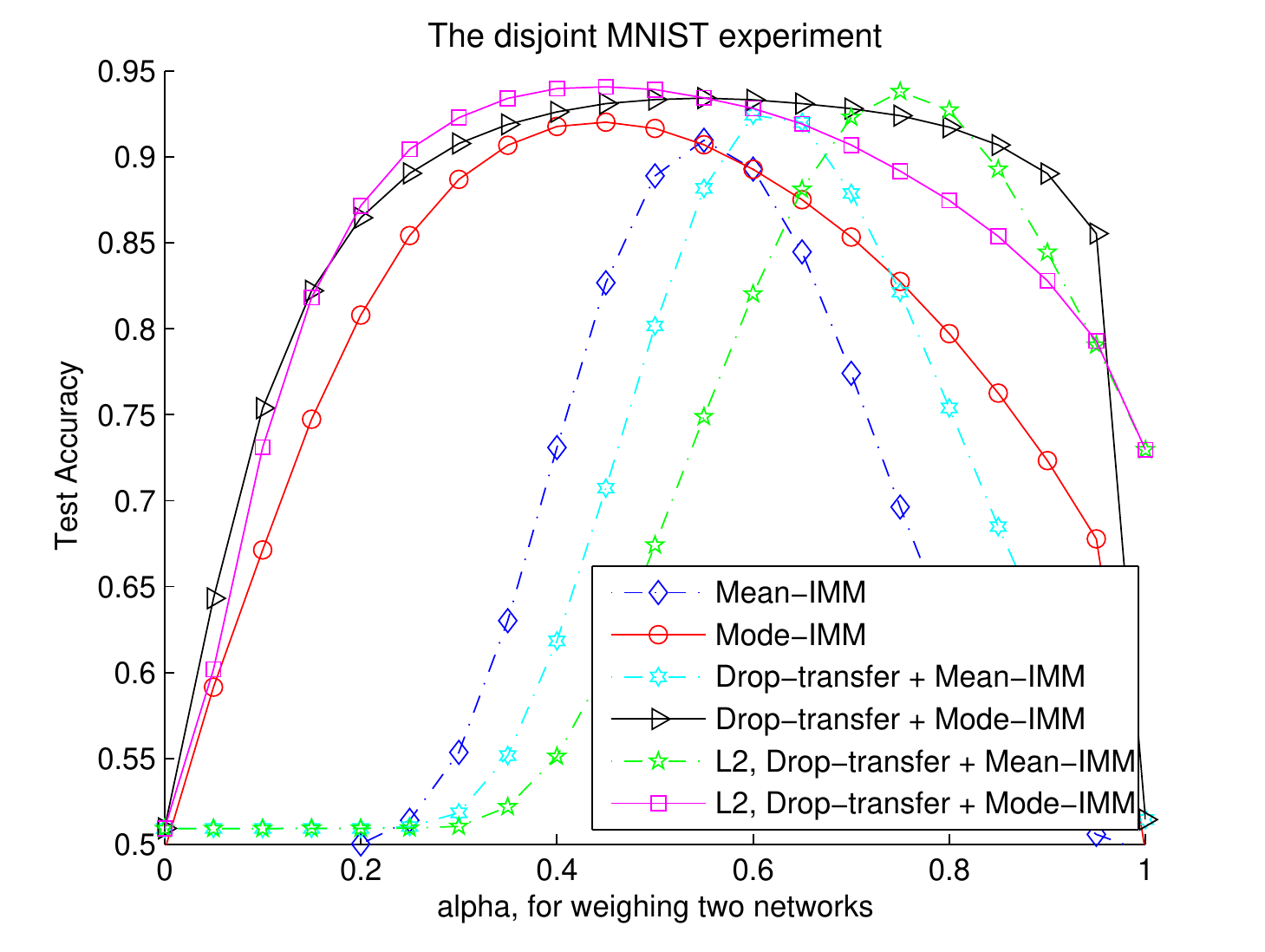}
\caption{\gt{Test accuracies of IMM with various transfer techniques on the disjoint MNIST. Both L2-transfer and drop-transfer boost the performance of IMM and make the optimal value of $\alpha$ larger than 1/2. However, drop-transfer tends to make the accuracy curve more smooth than L2-transfer does.}}
\label{fig:fig4}
\end{figure}

\gt{In our experimental setting, the usual SGD-based optimizers always perform less than 50\%, because the biases of the output layer for the old task are always pushed to large negative values, which implies that our task is} \rt{difficult.}
\gt{Figure \ref{fig:fig4} also shows that mode-IMM is robust with $\alpha$ and the optimal $\alpha$ of mean-IMM is larger than $1/2$ in the disjoint MNIST experiment.}

\textbf{Shuffled MNIST Experiment. }
The second experiment is the shuffled MNIST experiment \cite{goodfellow2013,kirkpatrick2016} of three sequential tasks.
In this experiment, the first dataset is the same as the original MNIST dataset.
However, in the second dataset, the input pixels of all images are shuffled with a fixed, random permutation. 
In previous work, EWC reaches the performance \ot{level} of the batch learner, and it is argued that EWC overcomes catastrophic forgetting in some domains.
The experimental details are similar to the disjoint MNIST experiment, except all models are allowed to use dropout regularization.
In the experiment, the first dataset is the same as the original MNIST dataset.
However, in the second and the third dataset, the input pixels of all images are shuffled with a fixed, random permutation\rt{, respectively.}
Therefore, the difficulty of the three datasets is the same, though a different solution is required for each dataset.

Table \ref{table:table1} (Bottom) and Figure \ref{fig:fig3} (Middle) shows the experimental results from the shuffled MNIST experiment.
Notice that accuracy of drop-transfer ($p$ = 0.2) alone is 96.86 ($\pm$ 0.21) and L2-transfer ($\lambda$ = 1e-4) + drop-transfer ($p$ = 0.4) alone is 97.61 ($\pm$ 0.15). These results are competitive to EWC without dropout, whose performance is around 97.0.

\textbf{ImageNet to CUB Dataset. }
The third experiment is the ImageNet2CUB experiment \cite{li2016}, the continual learning problem from the ImageNet dataset to the Caltech-UCSD Birds-200-2011 fine-grained classification (CUB) dataset \cite{wah2011}. 
The number\rt{s} of classes of ImageNet and CUB dataset \rt{are} around 1K and 200, and
the number\rt{s} of training instances \rt{are} 1M and 5K, respectively.
In the ImageNet2CUB experiment, the last-layer is separated for the ImageNet and the CUB task.
The structure of AlexNet is used for the trained model of ImageNet \cite{krizhevsky2012}.
\ot{In our experiment, we match the moments of the last-layer fine-tuning model and the LwF model,} with mean-IMM and mode-IMM.

Figure \ref{fig:fig3} (Right) shows that mean-IMM moderately balances the performance of two tasks between two networks. However, the balance\rt{d} \rt{hyperparameter} of mode-IMM is far from $\alpha$ = 0.5.
\rt{We think that} \ot{it is because} the scale of the Fisher matrix $F$ is different between the ImageNet and the CUB task.
\rt{Since} the number of training data of the two tasks is different, the mean of the square of the gradient, which is the definition of $F$, \rt{tends to be} different.
This implies that the assumption of mode-IMM does not always hold for \ot{heterogeneous tasks.}
See Appendix D.3 for more information including \rt{the} learning methods of IMM \rt{where} a different class output layer \rt{or} a different scale of \rt{the dataset is used}. 

Our results of IMM with LwF exceed the previous state-of-the-art performance, \ot{whose model is also LwF.}
\ot{This is because, in the previous works, the LwF model is initialized by the last-layer fine-tuning model, not directly by the original AlexNet.}
In this case, the performance loss of the old task is \rt{not only} decreased, but \rt{also} the performance gain of the new task is decreased.
The accuracies of our mean-IMM ($\alpha$ = 0.5) are 56.20 and 56.73 for the ImageNet task and the CUB task, respectively. The gains compared to the previous state-of-the-art are +1.13 and -1.14.
In the case of mean-IMM ($\alpha$ = 0.8) and mode-IMM ($\alpha$ = 0.99), the accuracies are 55.08 and 59.08 (+0.01, +1.12), and 55.10 and 59.12 (+0.02, +1.35), respectively.

\textbf{Lifelog Dataset. }
Lastly, we evaluate the proposed \rt{methods} on the Lifelog dataset \cite{lee2016}. The Lifelog dataset consists of 660,000 instances of egocentric video stream data, collected over 46 days from three participants using Google Glass \cite{lee2017}.
Three class categories, location, sub-location, and activity, are labeled on each frame of video.
In the Lifelog dataset, the class distribution changes continuously and new classes appear as the day passes.
Table \ref{table:table2} shows that mean-IMM and mode-IMM are competitive to the dual-memory architecture, the previous state-of-the-art ensemble model, even though IMM uses \rt{single} network.

\begin{table}[t]
\caption{Experimental results on the Lifelog dataset among different classes (location, sub-location, and activity) and different subjects (A, B, C). Every IMM uses weight-transfer. }
\centering
\small
\begin{tabular}{lccc C{1.15cm} C{1.15cm} C{1.15cm}} \hline
 & Location & Sub-location & Activity & A & B & C \\ \hline 
Dual memory architecture \cite{lee2016} 
& \textbf{78.11} & 72.36 & 52.92 & 67.02 & 58.80 & 77.57 \\ \hline
\textbf{Mean-IMM}  & 77.60 &  73.78 & 52.74 & 67.03 & 57.73 & \textbf{79.35} \\ 
\textbf{Mode-IMM}  & 77.14 & \textbf{75.76} & \textbf{54.07} & \textbf{67.97} & \textbf{60.12} & 78.89 \\ \hline
\label{table:table2}
\end{tabular}
\end{table}

\section{Discussion}

\textbf{A Shift of Optimal Hyperparameter of IMM. }
The tuned setting shows there often exists some $\alpha$ which makes the performance of the mean-IMM close to the mode-IMM.
However, in the untuned hyperparameter setting, mean-IMM performs worse when more transfer techniques are applied.
\rt{Our Bayesian interpretation in IMM assumes that the SGD training of the $k$-th network $\mu_k$ is mainly affected by the $k$-th task and is rarely affected by the information of the previous tasks.
However, transfer techniques break this assumption; thus the optimal $\alpha$ is shifted to larger than $1/k$.}
Fortunately, mode-IMM works more robustly than mean-IMM where transfer techniques are applied.
\rt{Figure \ref{fig:fig4} illustrates} the change of the test accuracy curve corresponding to the applied transfer techniques and \rt{the} following shift of the optimal $\alpha$ in mean-IMM and mode-IMM.

\textbf{Bayesian Approach on Continual Learning. }
Kirkpatrick et al. \cite{kirkpatrick2016} interpreted that the Fisher matrix $F$ as weight importance in explaining their EWC model.
In the shuffled MNIST experiment, since a large number of pixels always \rt{have} \ot{a} value of zero, the corresponding element\rt{s} of the Fisher matrix \rt{are} also zero.
Therefore, EWC does work by allowing weights to \ot{change, which} are not used in the previous task\rt{s}.
On the other hand, mode-IMM also works \ot{by selectively balancing between two weights using variance information.}
However, these assumptions on weight importance do not always hold, especially in the disjoint MNIST experiment.
The most important weight in the disjoint MNIST experiment is the bias term in the output layer. Nevertheless, these \rt{bias parts} of the Fisher matrix are not guaranteed to be the highest value nor can they be used to balance the class distribution between the first and second task.
We believe that using only the diagonal \rt{of} the covariance matrix in Bayesian neural networks is too na{\"i}ve in \ot{general and that this} is why EWC failed in the disjoint MNIST experiment.
We think \rt{it} could be alleviated in future work by using a more complex prior, such as a matrix Gaussian distribution \rt{considering the} correlations between nodes in the network \cite{Louizos2016}.

\textbf{Balancing the Information of an Old and a New Task. }
The IMM procedure produces a neural network without a performance loss for $k$th task $\mu_k$, which is better than the final solution $\mu_{1:k}$ in terms of the performance of the $k$th task.
Furthermore, IMM can easily weigh the importance of tasks in IMM models in real time.
For example, $\alpha_t$ can be easily changed for the solution of mean-IMM $\mu_{1:k} = \sum^k_t \alpha_t \mu_t$ .
\ot{In actual service situations of IT companies,} the importance of the old and the new task frequently changes in real time, and IMM can handle this problem.
This property differentiates IMM \rt{from the} other continual learning methods using the regularization approach, including LwF and EWC.


\section{Conclusion}
Our contributions are four fold\rt{s}.
First, we applied mean-IMM to the continual learning of \ot{modern} deep neural networks. Mean-IMM makes competitive results to comparative models and balances the information between an old and a new network.
We also interpreted the success of IMM by the Bayesian framework with Gaussian posterior. 
Second, we extended mean-IMM to mode-IMM with the interpretation of mode-finding \rt{in} the mixture of Gaussian posterior. Mode-IMM outperforms mean-IMM and comparative models in \rt{various} datasets.
Third, we introduced drop-transfer, a novel method \rt{proposed} in the paper.
Experimental results showed that drop-transfer alone performs well and is similar to the EWC without dropout, in the domain \rt{where} EWC rarely forgets.
Fourth, We applied various transfer techniques \rt{in} the IMM procedure to make our assumption of Gaussian distribution reasonable.
We argued that not only the search space of the loss function among neural networks can \ot{easily be} nearly convex, but also regularizers, such as dropout, make the search space smooth\rt{,} and the point in the search space have \rt{good accuracy.}
Experimental results showed that applying transfer techniques often boost the performance of IMM.
Overall, we made state-of-the-art performance in \rt{various} datasets of continual learning and explored geometrical properties and a Bayesian perspective \rt{of deep} neural networks.

\section*{\ot{Acknowledgments}}
The authors would like to thank Jiseob Kim, Min-Oh Heo, Donghyun Kwak, Insu Jeon, Christina Baek, and Heidi Tessmer for helpful comments and editing.
\rt{This work was supported by the Naver Corp. and partly by the Korean government (IITP-R0126-16-1072-SW.StarLab, IITP-2017-0-01772-VTT, KEIT-10044009-HRI.MESSI, KEIT-10060086-RISF). Byoung-Tak Zhang is the corresponding author.}

\bibliographystyle{unsrt}
\bibliography{mybibfile}

\begin{thebibliography}{10}

\bibitem{mccloskey1989}
Michael McCloskey and Neal~J Cohen.
\newblock Catastrophic interference in connectionist networks: The sequential
  learning problem.
\newblock {\em Psychology of learning and motivation}, 24:109--165, 1989.

\bibitem{french1999}
Robert~M French.
\newblock Catastrophic forgetting in connectionist networks.
\newblock {\em Trends in cognitive sciences}, 3(4):128--135, 1999.

\bibitem{goodfellow2013}
Ian~J Goodfellow, Mehdi Mirza, Da~Xiao, Aaron Courville, and Yoshua Bengio.
\newblock An empirical investigation of catastrophic forgetting in
  gradient-based neural networks.
\newblock {\em arXiv preprint arXiv:1312.6211}, 2013.

\bibitem{srivastava2013}
Rupesh~K Srivastava, Jonathan Masci, Sohrob Kazerounian, Faustino Gomez, and
  J{\"u}rgen Schmidhuber.
\newblock Compete to compute.
\newblock In {\em Advances in neural information processing systems}, pages
  2310--2318, 2013.

\bibitem{Ghahramani2000}
Zoubin Ghahramani.
\newblock Online variational bayesian learning.
\newblock In {\em NIPS workshop on Online Learning}, 2000.

\bibitem{Broderick2013}
Tamara Broderick, Nicholas Boyd, Andre Wibisono, Ashia~C Wilson, and Michael~I
  Jordan.
\newblock Streaming variational bayes.
\newblock In {\em Advances in Neural Information Processing Systems}, pages
  1727--1735, 2013.

\bibitem{li2016}
Zhizhong Li and Derek Hoiem.
\newblock Learning without forgetting.
\newblock In {\em European Conference on Computer Vision}, pages 614--629.
  Springer, 2016.

\bibitem{kirkpatrick2016}
James Kirkpatrick, Razvan Pascanu, Neil Rabinowitz, Joel Veness, Guillaume
  Desjardins, Andrei~A Rusu, Kieran Milan, John Quan, Tiago Ramalho, Agnieszka
  Grabska-Barwinska, et~al.
\newblock Overcoming catastrophic forgetting in neural networks.
\newblock {\em Proceedings of the National Academy of Sciences}, 2017.

\bibitem{mackay1992}
David~JC MacKay.
\newblock A practical bayesian framework for backpropagation networks.
\newblock {\em Neural computation}, 4(3):448--472, 1992.

\bibitem{Blundell2015}
Charles Blundell, Julien Cornebise, Koray Kavukcuoglu, and Daan Wierstra.
\newblock Weight uncertainty in neural network.
\newblock In {\em Proceedings of the 32nd International Conference on Machine
  Learning (ICML-15)}, pages 1613--1622, 2015.

\bibitem{goldberger2005}
Jacob Goldberger and Sam~T Roweis.
\newblock Hierarchical clustering of a mixture model.
\newblock In {\em Advances in Neural Information Processing Systems}, pages
  505--512, 2005.

\bibitem{lee2016}
Sang-Woo Lee, Chung-Yeon Lee, Dong~Hyun Kwak, Jiwon Kim, Jeonghee Kim, and
  Byoung-Tak Zhang.
\newblock Dual-memory deep learning architectures for lifelong learning of
  everyday human behaviors.
\newblock In {\em Twenty-Fifth International Joint Conference on Artificial
  Intelligencee}, pages 1669--1675, 2016.

\bibitem{rusu2016}
Andrei~A Rusu, Neil~C Rabinowitz, Guillaume Desjardins, Hubert Soyer, James
  Kirkpatrick, Koray Kavukcuoglu, Razvan Pascanu, and Raia Hadsell.
\newblock Progressive neural networks.
\newblock {\em arXiv preprint arXiv:1606.04671}, 2016.

\bibitem{fernando2017}
Chrisantha Fernando, Dylan Banarse, Charles Blundell, Yori Zwols, David Ha,
  Andrei~A Rusu, Alexander Pritzel, and Daan Wierstra.
\newblock Pathnet: Evolution channels gradient descent in super neural
  networks.
\newblock {\em arXiv preprint arXiv:1701.08734}, 2017.

\bibitem{huang2014}
Zhen Huang, Jinyu Li, Sabato~Marco Siniscalchi, I-Fan Chen, Chao Weng, and
  Chin-Hui Lee.
\newblock Feature space maximum a posteriori linear regression for adaptation
  of deep neural networks.
\newblock In {\em Fifteenth Annual Conference of the International Speech
  Communication Association}, 2014.

\bibitem{huang2015}
Zhen Huang, Sabato~Marco Siniscalchi, I-Fan Chen, Jinyu Li, Jiadong Wu, and
  Chin-Hui Lee.
\newblock Maximum a posteriori adaptation of network parameters in deep models.
\newblock In {\em Sixteenth Annual Conference of the International Speech
  Communication Association}, 2015.

\bibitem{rashwan2016}
Abdullah Rashwan, Han Zhao, and Pascal Poupart.
\newblock Online and distributed bayesian moment matching for parameter
  learning in sum-product networks.
\newblock In {\em Proceedings of the 19th International Conference on
  Artificial Intelligence and Statistics}, pages 1469--1477, 2016.

\bibitem{zhang10}
Kai Zhang and James~T Kwok.
\newblock Simplifying mixture models through function approximation.
\newblock {\em Neural Networks, IEEE Transactions on}, 21(4):644--658, 2010.

\bibitem{Pathak10}
Manas Pathak, Shantanu Rane, and Bhiksha Raj.
\newblock Multiparty differential privacy via aggregation of locally trained
  classifiers.
\newblock In {\em Advances in Neural Information Processing Systems}, pages
  1876--1884, 2010.

\bibitem{Baldi13}
Pierre Baldi and Peter~J Sadowski.
\newblock Understanding dropout.
\newblock In {\em Advances in Neural Information Processing Systems}, pages
  2814--2822, 2013.

\bibitem{ray2005}
Surajit Ray and Bruce~G Lindsay.
\newblock The topography of multivariate normal mixtures.
\newblock {\em Annals of Statistics}, pages 2042--2065, 2005.

\bibitem{pascanu2013}
Razvan Pascanu and Yoshua Bengio.
\newblock Revisiting natural gradient for deep networks.
\newblock {\em arXiv preprint arXiv:1301.3584}, 2013.

\bibitem{goodfellow2014}
Ian~J Goodfellow, Oriol Vinyals, and Andrew~M Saxe.
\newblock Qualitatively characterizing neural network optimization problems.
\newblock {\em arXiv preprint arXiv:1412.6544}, 2014.

\bibitem{yosinski2014}
Jason Yosinski, Jeff Clune, Yoshua Bengio, and Hod Lipson.
\newblock How transferable are features in deep neural networks?
\newblock In {\em Advances in neural information processing systems}, pages
  3320--3328, 2014.

\bibitem{evgeniou2004}
Theodoros Evgeniou and Massimiliano Pontil.
\newblock Regularized multi--task learning.
\newblock In {\em Proceedings of the tenth ACM SIGKDD international conference
  on Knowledge discovery and data mining}, pages 109--117. ACM, 2004.

\bibitem{kienzle2006}
Wolf Kienzle and Kumar Chellapilla.
\newblock Personalized handwriting recognition via biased regularization.
\newblock In {\em Proceedings of the 23rd international conference on Machine
  learning}, pages 457--464. ACM, 2006.

\bibitem{srivastava2014}
Nitish Srivastava, Geoffrey~E Hinton, Alex Krizhevsky, Ilya Sutskever, and
  Ruslan Salakhutdinov.
\newblock Dropout: a simple way to prevent neural networks from overfitting.
\newblock {\em Journal of Machine Learning Research}, 15(1):1929--1958, 2014.

\bibitem{wah2011}
Catherine Wah, Steve Branson, Peter Welinder, Pietro Perona, and Serge
  Belongie.
\newblock The caltech-ucsd birds-200-2011 dataset.
\newblock {\em Tech. Rep. CNS-TR-2011-001}, 2011.

\bibitem{krizhevsky2012}
Alex Krizhevsky, Ilya Sutskever, and Geoffrey~E Hinton.
\newblock Imagenet classification with deep convolutional neural networks.
\newblock In {\em Advances in neural information processing systems}, pages
  1097--1105, 2012.

\bibitem{lee2017}
Sang-Woo Lee, Chung-Yeon Lee, Dong-Hyun Kwak, Jung-Woo Ha, Jeonghee Kim, and
  Byoung-Tak Zhang.
\newblock Dual-memory neural networks for modeling cognitive activities of
  humans via wearable sensors.
\newblock {\em Neural Networks}, 2017.

\bibitem{Louizos2016}
Christos Louizos and Max Welling.
\newblock Structured and efficient variational deep learning with matrix
  gaussian posteriors.
\newblock {\em arXiv preprint arXiv:1603.04733}, 2016.

\bibitem{ray2012}
Surajit Ray and Dan Ren.
\newblock On the upper bound of the number of modes of a multivariate normal
  mixture.
\newblock {\em Journal of Multivariate Analysis}, 108:41--52, 2012.

\bibitem{amendola2017}
Carlos Am{\'e}ndola, Alexander Engstr{\"o}m, and Christian Haase.
\newblock Maximum number of modes of gaussian mixtures.
\newblock {\em arXiv preprint arXiv:1702.05066}, 2017.

\bibitem{kingma2013}
Diederik~P Kingma and Max Welling.
\newblock Auto-encoding variational bayes.
\newblock {\em arXiv preprint arXiv:1312.6114}, 2013.

\end{thebibliography}


\newpage
\onecolumn
\section*{APPENDIX A. Modes in the Mixture of Gaussian}
According to Ray and Lindsay \cite{ray2005}, all the critical points $\theta$ of a mixture of Gaussian (MoG) with two components are in one curve as the following equation with $0 < \alpha < 1$.

\begin{equation}
\theta = ((1-\alpha)\Sigma^{-1}_1 + \alpha \Sigma^{-1}_2)^{-1}
((1-\alpha)\Sigma^{-1}_1 \mu_1 + \alpha \Sigma^{-1}_2 \mu_2)
\label{eq:eq18}
\end{equation}

The proof is as follows. Imagine two Gaussian distribution $q_1$ and $q_2$, such as in Equation \ref{eq:eq2}.

\begin{gather}
q_1 \equiv q_1(\theta;\mu_1,\Sigma_1) = \frac{1}{\sqrt{(2\pi)^D|\Sigma_1|}}
\exp \left(-\frac{1}{2} (\theta - \mu_1)^T \Sigma^{-1}_1 (\theta - \mu_1) \right)\\
q_2 \equiv q_2(\theta;\mu_2,\Sigma_2) = \frac{1}{\sqrt{(2\pi)^D|\Sigma_2|}}
\exp \left(-\frac{1}{2} (\theta - \mu_2)^T \Sigma^{-1}_1 (\theta - \mu_2) \right)
\end{gather}

$D$ is the dimension of the Gaussian distribution.
Mixture of two Gaussian $q_1$ and $q_2$ with the equal mixing ratio (i.e., 1:1) is $q_1/2 + q_2/2$.
The derivation of the MoG is as follows:

\begin{equation}
\frac{\partial (q_1/2+q_2/2)}{\partial \theta} =
- \frac{q_1}{2} (\Sigma^{-1}_1 (\theta - \mu_1)) - \frac{q_2}{2} (\Sigma^{-1}_2 (\theta -\mu_2))=0
\label{eq:eq21}
\end{equation}

If we set Equation \ref{eq:eq21} to 0, to find all critical points, the following equation holds:

\begin{equation}
\theta = (q_1 \Sigma^{-1}_1 + q_2 \Sigma^{-1}_2)^{-1}
         (q_1 \Sigma^{-1}_1 \mu_1 + q_2 \Sigma^{-1}_2 \mu_2)
         \label{eq:eq22}
\end{equation}

When \rt{$\alpha$ is set} to $\frac{q_2}{q_1+q_2}$, Equation \ref{eq:eq18} holds.

Note that $\alpha_k$ is a function of $\theta$, so $\theta$ cannot be calculated in a closed-form from Equation \ref{eq:eq22}.
However, the optimal $\theta$ is in the set $\{\theta|\theta = ((1-\alpha) \Sigma^{-1}_1 + \alpha \Sigma^{-1}_2)^{-1} ((1-\alpha) \Sigma^{-1}_1 \mu_1 + \alpha \Sigma^{-1}_2 \mu_2), 0 < \alpha < 1\}$, which motivates our mode-IMM method.

In our study IMM uses diagonal covariance matrices, which means that there is no correlation between parameters. This diagonal assumption is useful, since it decreases the number of parameters for each covariance matrix from $O(D^2)$ to $O(D)$. Based on this, \rt{the $\theta$ in Equation \ref{eq:eq18}} is defined as follows:

\begin{equation}
    \rt{\theta_v = \frac{(1-\alpha) \cdot \mu_{1,v} / \sigma^2_{1,v} + \alpha \cdot \mu_{2,v} / \sigma^2_{2,v} }{ (1-\alpha)/\sigma^2_{1,v} + \alpha/\sigma^2_{2,v}}}
\end{equation}

$v$ denotes an index of the parameter vector. $\mu_{\cdot,v}$ and $\sigma^2_{\cdot,v}$ are scalar.

For MoG with two components in $K$ dimension, the number of modes can be at most $K+1$
\cite{ray2012}. Therefore, it is hard to find all modes in high-dimensional Gaussian in general.

The property of critical points of a MoG with two components can be extended to the case of $K$ components.
The following equation holds:
\begin{equation}
    \theta = (\sum^K_{k=1}\alpha_k \Sigma^{-1}_k)^{-1} (\sum^K_{k=1} \alpha_k \Sigma^{-1}_k \mu_k),
\end{equation}

where $0 < \alpha_k < 1$ for all $k$ and $\sum_k \alpha_k = 1$.
There is no tight upper bound on the number of modes of MoG in general.
There is a guess that, for all $D,K \geq 1$, the upper bound is \rt{$_{(D+K-1)}C_D$} \cite{amendola2017}.

\section*{APPENDIX B. \rt{Bayesian Neural Networks and} Continual Learning }


\textbf{Bayesian Neural Networks. }
Bayesian neural networks (BNN) assume an uncertainty for the whole parameter in neural networks so that the posterior distribution can be obtained \cite{Blundell2015}.
Previous studies have argued that BNN regularizes better than NN, and provides a confidence interval for the output estimation of each input instance.
Current research on BNN, to the best of our knowledge, uses Gaussian distributions as the posteriors of the parameters.
In the Gaussian assumption, because tracking the entire information of a covariance matrix is too expensive, researchers usually use only the diagonal term for the covariance matrix, \rt{where} the posterior distribution is fully factorized for each parameter.
However, the methods using full covariance \rt{were} also suggested recently \cite{Louizos2016}.
\rt{To estimate} a covariance matrix most studies use stochastic gradient variational Bayes (SGVB), where a sampled point from the posterior distribution by Monte Carlo is used in the training phases \cite{kingma2013}.
Alternatively, Kirkpatrick et al. \cite{kirkpatrick2016} approximated the covariance matrix as an inverse of a Fisher matrix.
This approximation makes the computational cost of the inference of a covariance matrix \rt{cheaper} when the update of covariance information is not needed in the training phase.
Our method follows the approach using the Fisher matrix.


\textbf{Elastic Weight Consolidation. } \rt{We compare} the work of Kirkpatrick et al. \cite{kirkpatrick2016} to the results of our framework.
The mechanism of EWC follows sequential Bayesian estimation.
EWC maximizes the following terms by gradient descent to get the solution $\mu_{1:K}$.

\begin{equation}
\begin{aligned}
\log p_{1:K} & \approx \log p(y_K|X_K,\theta) + \lambda \cdot \log p_{1:(K-1)} + C\\
& \approx \log p(y_K|X_K,\theta) + \lambda \cdot \sum^{K-1}_{k=1} \log q_{1:k} + C\\
& = \log p(y_K|X_K,\theta) - \frac{\lambda}{2} \cdot \sum^{K-1}_{k=1} (\theta - \mu_{1:k})^T \tilde{\Sigma}^{-1}_k (\theta - \mu_{1:k}) + C'
\label{eq:eq14}
\end{aligned}
\end{equation}

$p_k$ is empirical posterior distribution of $k$th task, and $q_k \sim N(\mu_k, \Sigma_k)$ is an approximation of $p_k$.
In EWC, $\tilde{\Sigma}^{-1}_k$ is also approximated by the diagonal term of Fisher matrix $\tilde{F}_k$ with respect to $\mu_{1:k}$ and $X_k$.

When moving to a third task, EWC uses the penalty term of both first and second network (i.e., $\mu_1$ and $\mu_{1:2}$).
Although this heuristic works reasonably in the experiments in their paper, it does not match to the philosophy of Bayesian.

\begin{algorithm}[t]
   \caption{IMM with weight-transfer, L2-transfer}
   \label{alg:alg1}
\begin{algorithmic}
   \STATE {\bfseries Input:} data \{$(X_1,y_1)$,$...$,$(X_K,y_K)$\}, balancing hyperparameter $\alpha$
   \STATE {\bfseries Output:} $w_{1:K}$
   \STATE $w_0 \leftarrow $ InitializeNN()
   \FOR {$k$ = 1:$K$}
       \STATE $w_{k*} \leftarrow w_{k-1}$
       \STATE Train$(w_{k*}, X_k, y_k)$ with $L(w_{k*}, X_k, y_k) + \lambda \cdot ||w_{k*} - w_{k-1}||^2_2$
       \IF{type is mean-IMM} 
          \STATE $w_{1:k} \leftarrow \sum^k_t \alpha_t w_{t*}$
       \ELSIF{type is mode-IMM}
          \STATE $F_{k*} \leftarrow$ CalculateFisherMatrix$(w_{k*}, X_k\rt{, y_k})$
          \STATE $\Sigma_{1:k} \leftarrow (\sum^k_t \alpha_t F_{t*})^{-1}$
          \STATE $w_{1:k} \leftarrow \Sigma_{1:k} \cdot (\sum^k_t \alpha_t F_{t*} w_{t*}) $
       \ENDIF
   \ENDFOR
\end{algorithmic}
\end{algorithm}

\textbf{Learning without Forgetting. } \rt{We compare} the work of Li and Hoiem \cite{li2016}. Although LwF does not explicitly assume Bayesian, the approach can be represented nonetheless as follows:
\begin{equation}
    \log p_{1:K} \approx \log p(y_K|X_K,\theta) + \lambda \cdot \sum^{K-1}_{k=1} \log p(\hat{y}_k|X_K,\theta)
\label{eq:eq15}
\end{equation}

Where $\hat{y_k}$ is the output from $\mu_k$ with input $X_K$. This framework is promising where the properties of a pseudo training set of $k$th task ($X_K$, $\hat{y}_k$) is similar to the ideal training set ($X_k$, $y_k$).

\section*{APPENDIX C. Example Algorithms of Incremental Moment Matching}

Two moment matching methods: mean-IMM and mode-IMM, and three transfer learning techniques: weight-transfer, L2-transfer, and drop-transfer, are combined to make \rt{various} continual learning algorithms in our study.
Algorithm \ref{alg:alg1} describes mean-IMM and mode-IMM with weight-transfer and L2-transfer.

\section*{APPENDIX D. Experimental Details}

Appendix D further explains following issues, 1) additional explanation of the untuned setting and tuned setting \rt{2)} techniques for IMM with a different class output layer for each task \rt{3)} other experimental details.

\subsection*{D.1 Disjoint MNIST Experiment}
We first explain the untuned setting and the tuned setting in detail.
The untuned setting refers to the \ot{most natural hyperparameter} in the equation of each algorithm, whereas the tuned setting refers to using heuristic hand-tuned hyperparameters.
For mean-IMM, it is \ot{most natural} to evenly average K models and $1/K$ is the most natural $\alpha_k$ value for $K$ sequential tasks.
For EWC, $1$ is the most natural $\lambda$ value in Equation \ref{eq:eq14}, because EWC is derived from the equation of sequential Bayesian.
For L2-transfer, there is no natural hyperparameter value in Equation \ref{eq:eq16}, so we need to heuristically choose a $\lambda$ value, which is not too small but does not damage the performance of the new network for the new task.

In the SGD, \rt{the number of epochs for the dataset (epoch per dataset)} for the second task corresponds to the hyperparameter. \ot{The unit} is how much of the network is trained from the whole data at once.
In the L2-transfer and EWC, $\lambda$ in Equations \ref{eq:eq16} and \ref{eq:eq14} corresponds to their \rt{hyperparameter}\ot{.}
In the mean-IMM and mode-IMM, $\alpha_K$ in Equations \ref{eq:eq5} and \ref{eq:eq9} corresponds to the \rt{hyperparameter}\ot{.}
In the drop-transfer, dropout ratio $p$ in Equation \ref{eq:eq17} corresponds to the hyperparameter.

All of the explained hyperparameters are devised to balance the information between the old and new tasks.
If $\lambda/(1+\lambda) = 1$ or $\alpha_1 = 1$, the final network of the algorithms is the same as the network for the first task.
If $1/(1+\lambda) = 1$ or $\alpha_K = 1$, the final network is the same as the network for the last task.

We used multi-layer perceptrons (MLP) with \rt{[784-800-800-10]} as the number of nodes, ReLU as the activation function, and vanilla SGD as the optimizer for all of the experiments.
We set \ot{the epoch per dataset} to 10, \ot{unless} otherwise noted.
The entire IMM model uses weight-transfer to smooth the loss surface of the model. Without weight-transfer, our IMM model does not work at all.
In our experiments, all models only use one 10-way softmax output layer.
\rt{For} only SGD, dropout is used as proposed in Goodfellow et al. \cite{goodfellow2013}, but dropout does not help much.

Each accuracy was measured by \rt{averaging the results of} 10 experiments.
In the experiment, IMM outperforms comparative models by a significant margin.
In the tuned experiment, the performance of the IMM models exceeds 90\%, and the performance \ot{increases} more when more transfer techniques are applied.
Among all the models, weight-transfer + L2-transfer + drop-transfer + mode-IMM performs the best and \rt{its performance is} greater than 94\%.
However, the comparative models fail to reach greater than 90\%.
Existing regularizer including dropout does not improve the comparative models.

\begin{figure}[t] 
\centering
\includegraphics[width=0.45\textwidth]{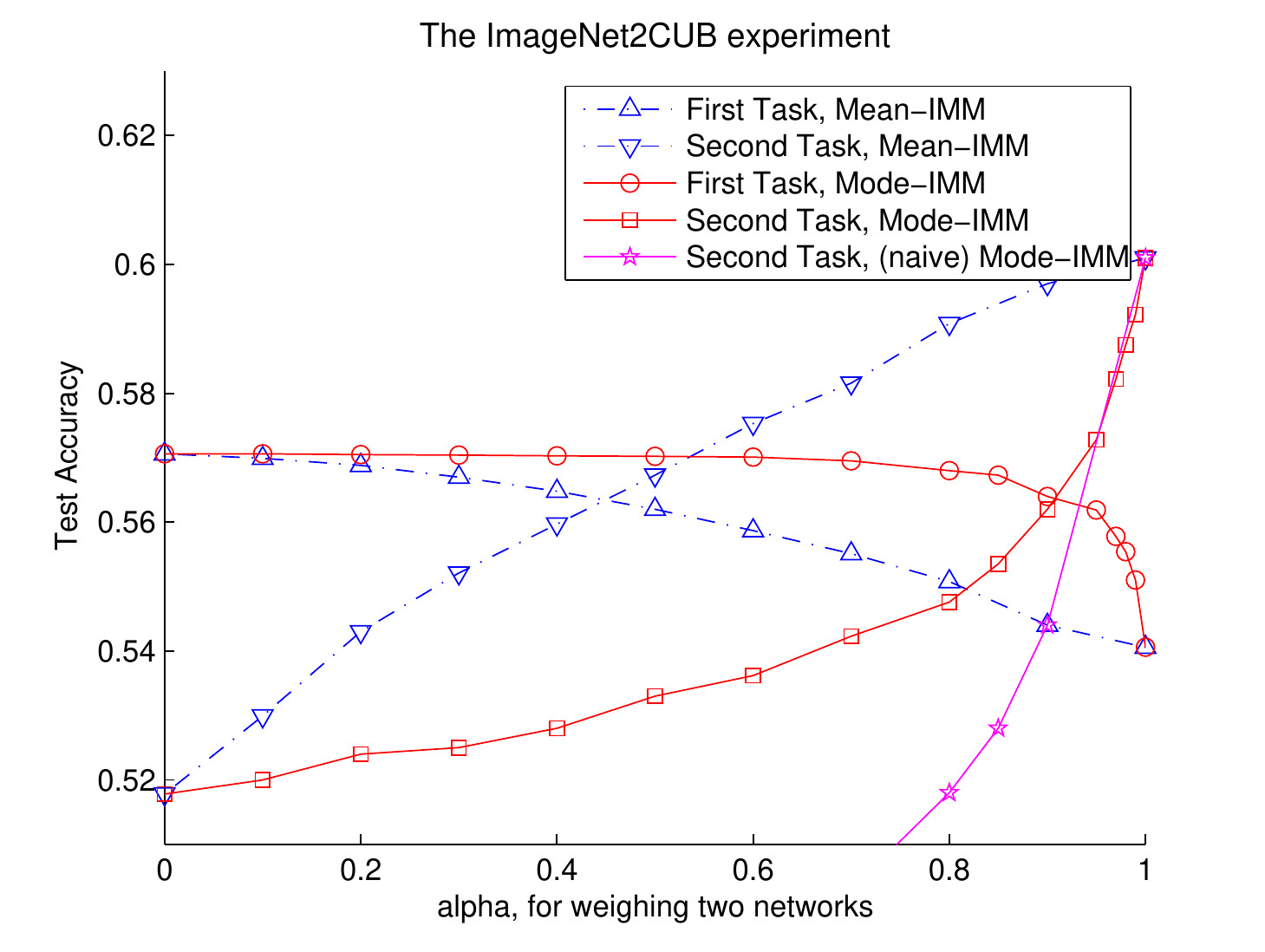}
\includegraphics[width=0.45\textwidth]{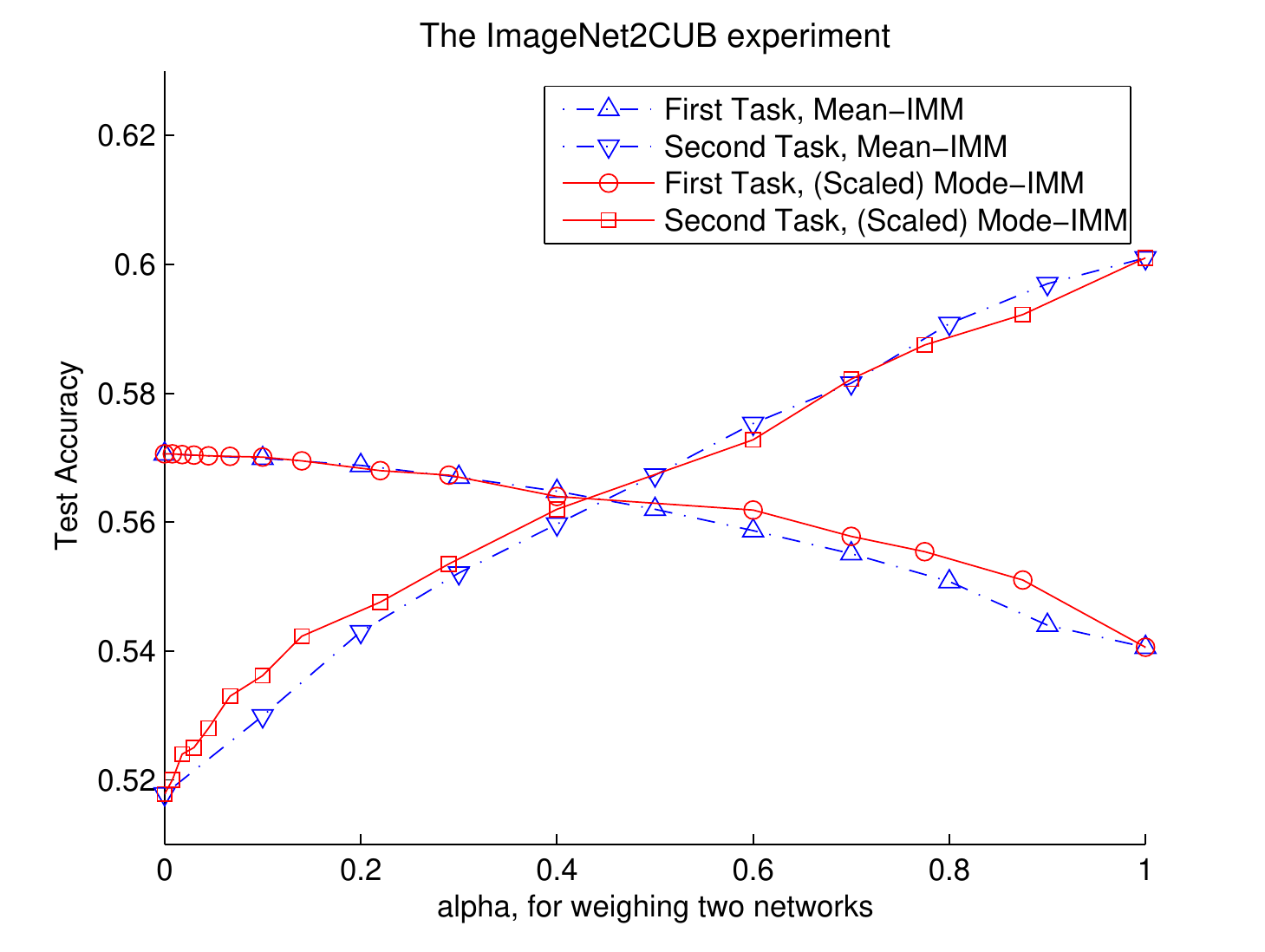}
\caption{(Left) Illustration of the effect of the strategy of re-weighing on the new last-layer. Mode-IMM refers to the original mode-IMM devised for the ImageNet2CUB experiments. In na{\"i}ve mode-IMM, the second last-layer of the second network is used for the second last-layer of the final IMM model. 
(Right) The results of mode-IMM with changing the balancing hyperparameter $\alpha$ to the re-scaled balancing hayperparameter $\hat{\alpha}$ with the scale of the Fisher matrix of each network.}
\label{fig:fig5}
\end{figure}

\subsection*{D.2 Shuffled MNIST Experiment}

The second experiment is the shuffled MNIST experiment for three sequential tasks.
For the hyperparameter of IMM, we set $\alpha_1$ and $\alpha_2$ as the same value, and tune only $\alpha_3$.
Table \ref{table:table1} (Bottom) shows the experimental results. The performance of SGD + dropout and EWC + dropout comes from the report in \cite{kirkpatrick2016}. Changing only the epoch does not \rt{significantly} increase the performance in SGD.  
\rt{The results} show that our IMM \rt{paradigm performs} similarly to EWC \rt{in a case where  EWC performs well.}
Dropout regularization in the task makes both our models and comparative models perform better.

In our IMM framework, weight-transfer, L2-transfer, and drop-transfer all \rt{take} $\mu_{k-1}$ as the reference models of the transfer for training $\mu_k$.
In other words, weight-transfer initializes $\mu_k$ with $\mu_{k-1}$, L2-transfer uses a regularization term to minimize the Euclidean distance between $\mu_{k-1}$ and $\mu_k$, and drop-transfer uses a $\mu_{k-1}$ as the zero point of the dropout procedure.
All three transfer techniques can be considered to change the reference point to, for example, $\mu^{mean}_{1:(k-1)}$ or $\mu^{mode}_{1:(k-1)}$, as previous works do \cite{kirkpatrick2016}.
However, \rt{all these alternatives make performances worse} in our shuffled MNIST experiment.
We argued that our utilization of transfer techniques \rt{is} devised not to minimize the distance between $\mu_{k-1}$ and $\mu_k$, but to help find a $\mu_k$ with a smooth and convex-like loss space between $\mu_{k-1}$ and \rt{$\mu_k$.}


\subsection*{D.3 ImageNet to Other Image Datasets}

\rt{When each task needs a different class output layer}, IMM requires additional techniques.
There is no counterpart \rt{weight matrix} in the last-layer of the first network representing the second task, \rt{nor} the second last-layer of the first network. 
To tackle this problem, we add the training process of the last-layer fine-tuning model to the IMM procedure; we match the moments of the last-layer fine-tuning model \rt{with} the original new network for the new task. 
Last-layer fine-tuning is the model the last-layer is only fine-tuned for each new task; thus it does not make a performance loss for the first task, but does not often learn enough for new tasks.

The technique utilizing the last-layer fine-tuning model makes mean-IMM work in the case of different class output layers, but it is not enough for mode-IMM.
It is not possible to calculate a proper Fisher matrix of the second last-layer in the first network for the first dataset.
As the Fisher matrix is defined with the gradient from the loss of the first task, elements of the Fisher matrix have a value of zero.
However, a zero matrix not only is what we \rt{do not} want but also degenerates the performance of mode-IMM.
To tackle this problem, we apply mean-IMM for the last-layer with a re-scaling.
We change the mixing ratios $\alpha_1$ : $\alpha_2$ to $\hat{\alpha}_1$ : $\hat{\alpha}_2$ = $\alpha_1$ : $\alpha_2 \cdot \frac{|\hat{w}_1|}{|\hat{w}_1|+|\hat{w}_2|}$ for the re-scaling, 
where $|\hat{w}_1|$ and $|\hat{w}_2|$ is the average of the whole element of weight matrix in the layer before the last-layer, in the first and the second task.

In our ImageNet2CUB experiment, the moments of the last-layer fine-tuning model and the LwF model are matched.
Though LwF does not perform well in our previous experiments, \ot{it is known that} LwF performs well when the size of a new dataset is small relative to the old dataset, \rt{as in} the ImageNet2CUB experiment.

Figure \ref{fig:fig5} (Left) compares the performances of mode-IMM models with different assumptions on the Fisher matrix.
In na{\"i}ve mode-IMM, the Fisher matrix of the second last-layer of the first network is a zero matrix. In other words, the second last-layer of the final na{\"i}ve mode-IMM is the second last-layer of the second network.
Na{\"i}ve mode-IMM does not yield a good performance as we expect.

In Figure \ref{fig:fig5}, scaled mode-IMM denotes the results of mode-IMM re-plotted by the $\hat{\alpha}$ as we defined above. The result shows that re-scaled mode-IMM performs similarly to mean-IMM in the ImageNet2CUB experiment.

\subsection*{D.4 Lifelog Dataset}

\begin{table}[t]
\caption{Experimental results on the Lifelog dataset. Mean-IMM uses weight-transfer.  Classification accuracies among different classes (Top) and different subjects (Bottom).
In the experiment, our IMM paradigm achieves competitive results with the approach using an ensemble network, without additional cost for inference and learning.}
\centering
\small
\begin{tabular}{lccc} \\\hline
Algorithm & Location & Sub-location & Activity \\ \hline \hline
Dual memory architecture \cite{lee2016} & \textbf{78.11} & 72.36 & 52.92 \\ \hline
\textbf{Mean-IMM}  & 77.60 &  73.78 & 52.74 \\ \hline 
\textbf{Mode-IMM}  & 77.14 & \textbf{75.76} & \textbf{54.07} \\ \hline \hline
Online fine-tuning & 68.27 & 64.13  & 50.00  \\ \hline
Last-layer fine-tuning & 74.58 & 69.30 & 52.22 \\ \hline
Na{\"i}ve incremental bagging & 74.48 & 67.18 & 47.92 \\ \hline
Incremental bagging w/ transfer & 74.95 & 68.53  & 49.66  \\ \hline
\\ \hline
Algorithm & A & B & C \\ \hline \hline
Dual memory architecture \cite{lee2016} & 67.02 & 58.80 & 77.57 \\ \hline
\textbf{Mean-IMM} & 67.03 & 57.73 & \textbf{79.35} \\ \hline 
\textbf{Mode-IMM} & \textbf{67.97} & \textbf{60.12} & 78.89 \\ \hline \hline
Online fine-tuning & 53.01 & 56.54 & 72.85  \\ \hline
Last-layer fine-tuning & 63.31 & 55.83 & 76.97 \\ \hline
Na{\"i}ve incremental bagging & 62.24 & 53.57 & 73.77 \\ \hline
Incremental bagging w/ transfer & 61.21 & 56.71  & 75.23  \\ \hline
\label{table:table3}
\end{tabular}
\end{table}

The Lifelog dataset is the dataset recorded from Google Glass over 46 days from three participants. 
The 660,000 seconds of the egocentric video stream data reflects the behaviors of the participants.
The dataset consists of 10 days of training data and 4 days of test data in order of time for each participant respectively.
In the framework of Lee et al. \cite{lee2016}, the network can be updated every day, but a new network can be made for the 3rd, 7th, and 10th day, with training data of 3, 4, and 3 days, respectively.
Following this framework, our network is made in the 3rd, 7th, and 10th day, and then merged to previously trained networks.
Our IMM used AlexNet pretrained by the ImageNet dataset \cite{krizhevsky2012} as the initial network. 
The experimental results on the Lifelog dataset are in Table \ref{table:table3}, where the performance of models is from Lee et al. \cite{lee2016} except IMM.

\end{document}